\documentclass[]{interact}

\usepackage{epstopdf}
\usepackage{subfigure}

\usepackage{natbib}
\bibpunct[, ]{(}{)}{;}{a}{}{,}

\theoremstyle{plain}

\theoremstyle{definition}

\theoremstyle{remark}

\usepackage{hyperref}
\usepackage[utf8]{inputenc}

\usepackage{caption}
\usepackage{graphicx}
\usepackage{xcolor}
\usepackage[ruled,vlined]{algorithm2e}
\usepackage{booktabs}
\usepackage{longtable}
\usepackage{array}
\usepackage{multirow}
\usepackage{wrapfig}
\usepackage{float}
\usepackage{colortbl}
\usepackage{pdflscape}
\usepackage{tabu}
\usepackage{threeparttable}
\usepackage{threeparttablex}
\usepackage[normalem]{ulem}
\usepackage{makecell}
\usepackage{setspace}

\begin{document}

\articletype{ARTICLE}

\title{Neural Networks for Latent Budget Analysis of Compositional Data}

\author{\name{Zhenwei Yang$^{a}$, Ayoub Bagheri and P.G.M van der Heijden}
\affil{$^{a}$Utrecht University, Sjoerd Groenman building, Padualaan 14,
3584 CH, Utrecht, The Netherlands}
}

\thanks{CONTACT Zhenwei
Yang. Email: \href{mailto:z.yang@uu.nl}{\nolinkurl{z.yang@uu.nl}}}

\maketitle

\begin{abstract}
Compositional data are non-negative data collected in a rectangular
matrix with a constant row sum. Due to the non-negativity the focus is
on conditional proportions that add up to 1 for each row. A row of
conditional proportions is called an observed budget. Latent budget
analysis (LBA) assumes a mixture of latent budgets that explains the
observed budgets. LBA is usually fitted to a contingency table, where
the rows are levels of one or more explanatory variables and the columns
the levels of a response variable. In prospective studies, there is only
knowledge about the explanatory variables of individuals and interest
goes out to predicting the response variable. Thus, a form of LBA is
needed that has the functionality of prediction. Previous studies
proposed a constrained neural network (NN) extension of LBA that was
hampered by an unsatisfying prediction ability. Here we propose LBA-NN,
a feed forward NN model that yields a similar interpretation to LBA but
equips LBA with a better ability of prediction. A stable and plausible
interpretation of LBA-NN is obtained through the use of importance plots
and table, that show the relative importance of all explanatory
variables on the response variable. An LBA-NN-K-means approach that
applies K-means clustering on the importance table is used to produce K
clusters that are comparable to K latent budgets in LBA. Here we provide
different experiments where LBA-NN is implemented and compared with LBA.
In our analysis, LBA-NN outperforms LBA in prediction in terms of
accuracy, specificity, recall and mean square error. We provide
open-source software at GitHub.
\end{abstract}


\hypertarget{introduction}{%
\section{Introduction}\label{introduction}}

Compositional data are non-negative constant-row-sum data in a
rectangular matrix. Due to this non-negativity and constant-row-sum,
interest usually goes out to, for each row in the matrix, relative
quantities for the different column categories that add up to one
\citep{PG2006}. Compositional data are present in many fields, such as
archaeology, social science and medical science
\citep{Aitchison1994, Aitchison2008, baxter2006, PG2017, raun2016}.
Often the rows represent individuals, or groups of individuals, and the
columns levels of a response variable. As an example, one can think of
several groups of people from different continents choosing different
ways of payment, where for each group the number of individuals is
subdivided over the ways of payment. The non-negativity and the constant
row sum distinguish compositional data from other types of data.
Therefore, classical statistical techniques that rely on product-moment
covariance and correlation are poorly suited for compositional data
\citep{Aitchison1994, Aitchison2008}.

Latent budget analysis (LBA) \citep{Heijden1992} is one of the
statistical methods to study compositional data. A row of conditional
proportions in the rectangular matrix is called an observed budget. LBA
approximates the observed budgets with a mixture of latent budgets,
where the latent budgets and the mixture weights are estimated from the
data. LBA is usually applied to a contingency table with \(I\) groups
(levels of the categorical explanatory variable) and \(J\) levels of the
response variable. LBA allows to study the dependence of the response
variable on the categorical explanatory variable(s). However, in
prospective studies, knowledge of the \(J\) categories of the response
variable is usually missing and there is only information on the
categorical explanatory variable(s). A challenge of predicting the \(J\)
categories based on \(I\) groups is imposed. For this purpose, the
traditional LBA is restricted as it lacks the ability of predicting the
response variable.

In previous work, \cite{SM2001} extended LBA with a constrained neural
network (NN) model, which in the rest of the article is called the
constrained NN extension of LBA. Their method allows to use LBA as a
prediction model. However, the model performance in prediction was not
explicitly discussed there. We argue that using the constrained NN
extension can hamper high accuracy in prediction. Here we propose to use
unconstrained NN techniques to analyse compositional data and provide
add-ons for LBA. Two research questions are answered: (1) How can we
equip LBA with the ability to predict the response variable based on the
explanatory variable(s) in prospective studies? (2) Can we achieve a
better prediction ability with the new, unconstrained NN-based version
of LBA than with the constrained NN extension of LBA?

In the remainder of this paper, in Section \ref{sec:bg}, we explain LBA
and the extension proposed by Siciliano and Mooijaart. In Section
\ref{sec:method}, an unconstrained NN extension of LBA, LBA-NN, is
proposed for analysing and predicting compositional data. We shed light
on using the importance table and plot that show the relative importance
of all explanatory variables on the response variable to obtain a stable
interpretation of LBA-NN. The LBA-NN-K-means approach that applies
K-means clustering on the importance table is proposed for LBA-NN as a
qualitative comparison with LBA. Besides, we compare LBA and LBA-NN
quantitatively based on six standard measurements of the prediction
ability. We provide an R function for LBA-NN at GitHub
\footnote{\label{footnote:lbann} \url{https://github.com/ZhenweiYang96/lbann}}.
In Section \ref{sec:app}, we apply LBA-NN on different example datasets.
In Section \ref{sec:dis}, a brief discussion about LBA-NN and the
expectation of the potential usage is shown.

\hypertarget{sec:bg}{%
\section{Background}\label{sec:bg}}

During the past several decades, many models have been proposed that aim
to provide a better understanding of compositional data
\citep{desarbo1993, hsu2019}. Here we focus on LBA. In this section, the
previous studies on LBA are reviewed. In Section \ref{subsection:lba},
LBA is explained. In Section \ref{subsection:prelba}, the preliminary
constrained NN extension of LBA is demonstrated.

\hypertarget{subsection:lba}{%
\subsection{Latent budget analysis}\label{subsection:lba}}

LBA is a mixture model to study compositional data without relying on a
covariance matrix. LBA was introduced and coined by \cite{devan1988}.
Earlier work can be found in \cite{goodman1974}.

Imagine an \(I \times J\) matrix \(P\), the rows are \(I\) levels of the
explanatory variable(s) and the columns \(J\) levels of the response
variable. Each row \(i\) is regarded as an observed budget (denoted by
\(\bm{p}_i\)) with \(J\) observed levels \(p_{j|i}\). To detect the
relationship between the explanatory variable(s) and the response
variable, \(K\) latent budgets (\(K \leq min(I,J)\)) are introduced,
denoted by \(\boldsymbol\beta_k\). Each observed budget \(\bm{p}_i\) is
approximated by a mixture \(\boldsymbol\pi_i\), of latent budgets
\(\boldsymbol\beta_k\), in the followinng way:

\begin{equation}
\label{eqn:11}
\boldsymbol\pi_i = \alpha_{1|i}\boldsymbol\beta_1 + \cdots + \alpha_{k|i}\boldsymbol\beta_k + \cdots + \alpha_{K|i}\boldsymbol\beta_K \tag{1.1}
\end{equation} where \(\alpha_{k|i}\) \((i=1,\ldots,I;k=1, \ldots,K)\)
are the mixing parameters adding up to 1 for each \(i\). Different
contributions from the latent budgets are revealed by the mixing
parameters for all levels of the explanatory variable(s). The elements
in the expected budgets are denoted by \(\pi_{j|i}\), which are the
counterparts of observed \(p_{j|i}\). Formula \ref{eqn:11} can be
rewritten as: \begin{equation}
  \pi_{j|i} = \sum^K_{k = 1}{\alpha_{k|i} \beta_{j|k}} \tag{1.2}
\end{equation} where \(\beta_{j|k} (j=1,\ldots,J;k=1, \ldots,K)\) are
the elements of the latent budgets. LBA can also be expressed in matrix
notation: \begin{equation}
\Pi = AB^T \tag{1.3}
\end{equation} where \(\Pi\) is an \(I \times J\) matrix whose rows are
expected budgets; \(A\) is an \(I \times K\) matrix of mixing
parameters; \(B\) is a \(J \times K\) matrix whose columns are latent
budgets. These parameter matrices are subject to the properties of
compositional data, sum-to-constant: \begin{equation}
\sum^I_{i=1}{\pi_{j|i}} = \sum^K_{k=1}{\alpha_{k|i}} = \sum^J_{j=1}{\beta_{j|k}} = 1 \tag{1.4}
\end{equation} and non-negativity: \begin{equation}
0 \leq \pi_{j|i} \leq 1; 0 \leq \alpha_{k|i} \leq 1; 0 \leq \beta_{j|k} \leq 1 \tag{1.5}
\end{equation}

\par

\begin{figure}[H]
\subfigure[\label{fig:1a}]{\includegraphics[width = 7cm]{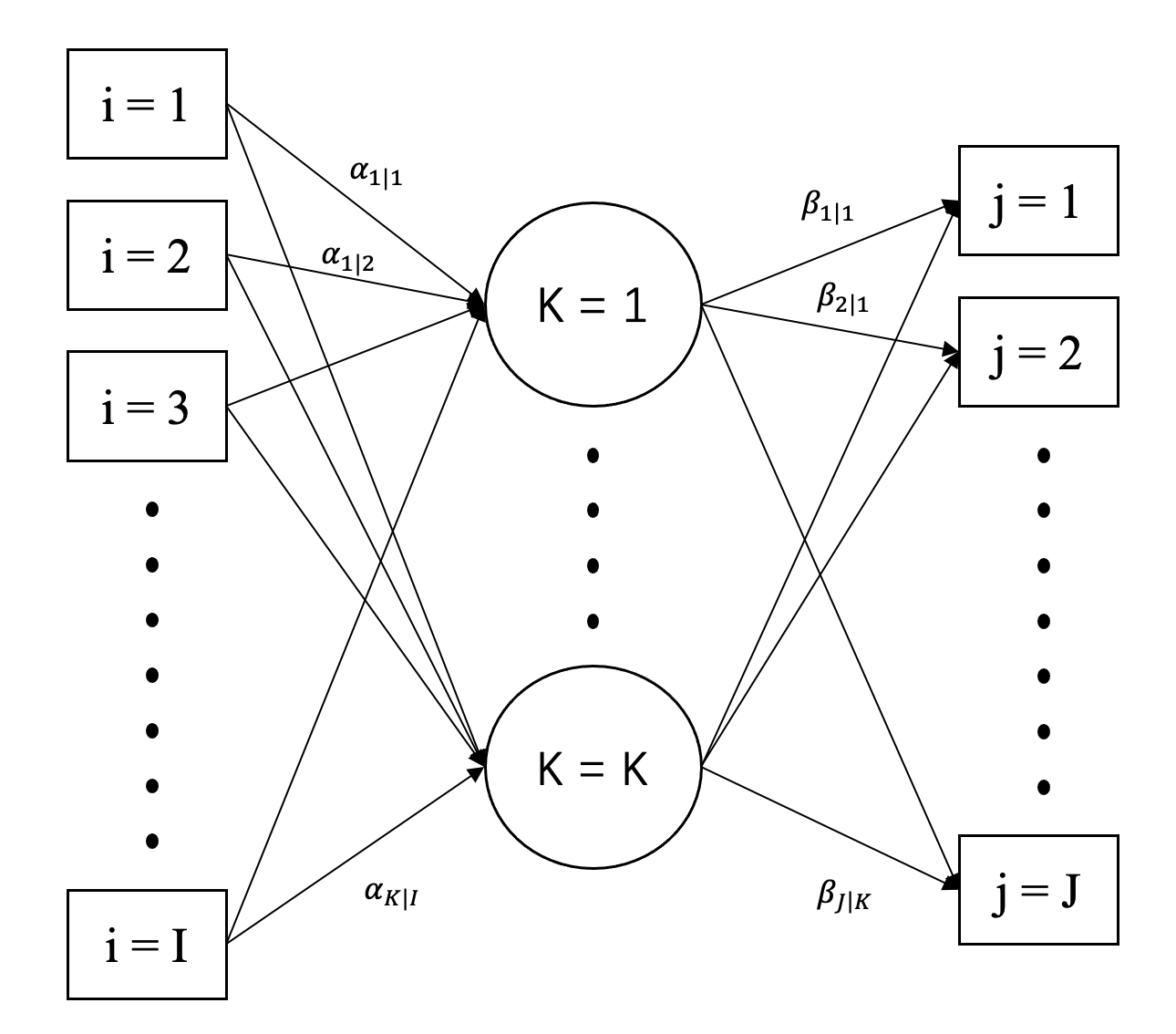}}
\subfigure[\label{fig:1b}]{\includegraphics[width=7cm]{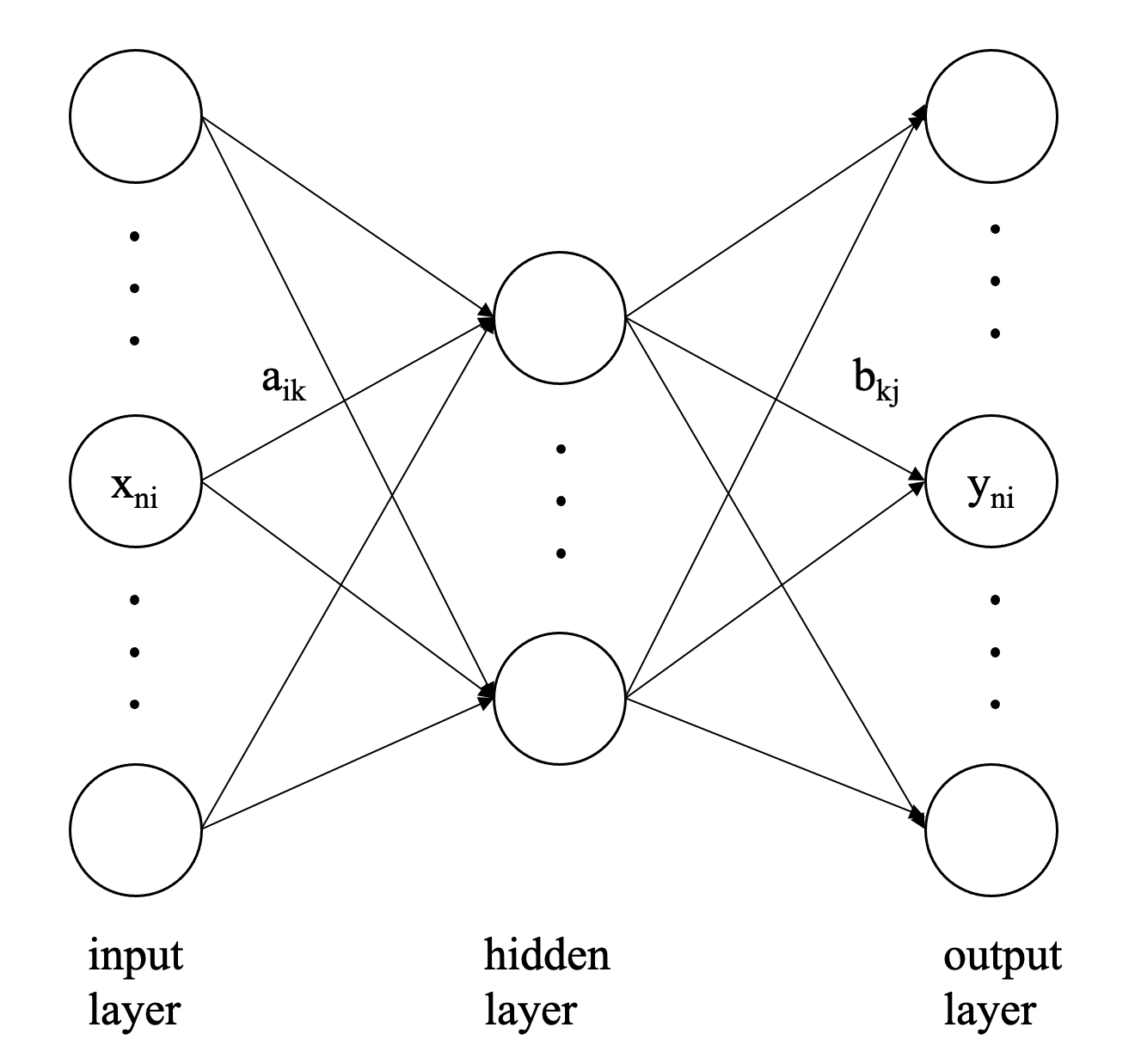}}
\caption{The graphical display of LBA and its constrained extension. (a: the MIMIC model interpretation of LBA; b: the constrained extension of LBA by Siciliano and Mooijaart (2001))}
\end{figure}

One way to interpret the model is called the Multiple Indicator Multiple
Cause (MIMIC) model interpretation \citep{vanderark1998}. The graphical
presentation of the MIMIC model is displayed in Figure \ref{fig:1a}.
Here, it is assumed that the explanatory variable(s) and response
variable are independent given the \(K\) latent budgets. The model
parameters are conditional probabilities. For example, \(\alpha_{k|i}\)
is delineated as the probability that the \(k^{th}\) latent budget
covers the characteristics of the \(i^{th}\) category of the explanatory
variables; \(\beta_{j|k}\) denotes the probability of the \(j^{th}\)
category in the \(k^{th}\) latent budget; \(\pi_{j|i}\) is the
probability that one falls in the \(j^{th}\) category of the response
variable if one falls in the \(i^{th}\) category of the explanatory
variable.

An example of data on which LBA can be applied is shown in Table
\ref{tab:example_lba}. The rows represent the continent where the
individuals live and the columns are the ways of payment. In this
scenario, LBA can assist to find the relation between the continents and
ways of payment through introducing \(K\) latent budgets. LBA finds
latent budgets that stand for typical profiles of payment, for example,
when \(K = 2\), for \(k = 1\), an overrepresentation of more modern ways
of payment and, for \(k = 2\), an overrepresentation of more
old-fashioned ways of payment. The mixing parameters reveal whether a
continent makes more use of the modern or the old-fashioned ways of
payment.

\begin{table}[H]

\caption{\label{tab:example_lba}An example contingency table of the scenario to apply LBA}
\centering
\begin{tabular}[t]{lcccc}
\toprule
\textbf{ } & \textbf{Cash} & \textbf{Credit Cards} & \textbf{Mobile App} & \textbf{Check}\\
\midrule
America & 0.07 & 0.70 & 0.20 & 0.03\\
Asia & 0.25 & 0.10 & 0.60 & 0.05\\
Europe & 0.35 & 0.45 & 0.15 & 0.05\\
Africa & 0.80 & 0.10 & 0.08 & 0.02\\
Oceania & 0.05 & 0.40 & 0.35 & 0.20\\
\addlinespace
Other & 0.60 & 0.15 & 0.15 & 0.10\\
\bottomrule
\end{tabular}
\end{table}

\hypertarget{subsection:prelba}{%
\subsection{A preliminary constrained NN extension of
LBA}\label{subsection:prelba}}

A multi-layer perceptron neural network model refers to a type of
feedforward neural network, including an input layer, one or more hidden
layers and an output layer. Each piece of data from one subject passes
through the layers in a one-directional way. Each layer is comprises one
or more neurons. Different layers are connected by assigned weights
between the neurons.

The MIMIC model interpretation (see Section \ref{subsection:lba})
resembles the structure of NN. Thus, \cite{SM2001} proposed a different
way of interpreting LBA as a supervised double-layer perceptron neural
network. A graphical display is presented in Figure \ref{fig:1b}, where
the neurons are displayed by circles. Take the example in Section
\ref{subsection:lba}, where the input layer consists of five categories
of the explanatory variable, ``continent''. The neurons of the hidden
layer are two latent budgets and the output layer contains four
categories of the response variable, ``ways of payment''. The matrix
\(A\) of mixing parameters in LBA is interpreted as the input-hidden
weight matrix in NN, constrained with constant row sum and
non-negativity. Similarly, the matrix \(B\) of latent budgets is
interpreted as the transpose of the hidden-output weight matrix,
constrained with constant column sum and non-negativity. In that way,
the formula for LBA can be rewritten by: \begin{align*}
\label{equ:siciliano}
    Y & = \sigma_2(\sigma_1(XA + b_1)B^T + b_2) + E \ \ \ \ \mathrm{s.t.} \ \sigma_1(x) = \sigma_2(x); b_1 = 0_{N,K}, b_2 = 0_{N,J} \\
    & = XAB^T + E \tag{2}
\end{align*} where \(X\) is an \(N \times I\) input data matrix with
dummy variables, \(N\) is the sample size and \(I\) is the number of
categories of the explanatory variable(s); likewise, \(Y\) is an
\(N \times J\) output data matrix with dummy variables where \(J\) is
the number of categories of the response variable; \(\sigma_1(x)\) and
\(\sigma_2(x)\) are called the identity transfer functions here; \(b_1\)
and \(b_2\) are the bias term in the hidden layer and output layer,
respectively, restricted as zero matrices. With the bias terms
restricted as zero matrices, residuals are included in the error term,
\(E\).

\hypertarget{sec:method}{%
\section{Proposed methods}\label{sec:method}}

To improve the prediction ability of the extension of LBA by Siciliano
and Mooijaart, we propose an NN extension where the number of hidden
neurons can be larger than K and the weight matrices are unconstrained,
i.e., matrix \(A\) without constraints of constant row sum and
non-negativity, and matrix \(B\) without constraints of constant column
sum and non-negativity. We elaborate the model in Section
\ref{subsection:lbann}. We also provide an R function for LBA-NN to
facilitate the implementation in Section \ref{subsection:function}. An
interpretation strategy is described in Section
\ref{subsection:interpret}. The quantitative comparison strategy to
compare LBA and LBA-NN is introduced in Section \ref{subsection:quanti}.

\hypertarget{subsection:lbann}{%
\subsection{LBA-NN: Latent Budget Analysis - Neural
Network}\label{subsection:lbann}}

\begin{figure}[H]

{\centering \includegraphics[width=10cm]{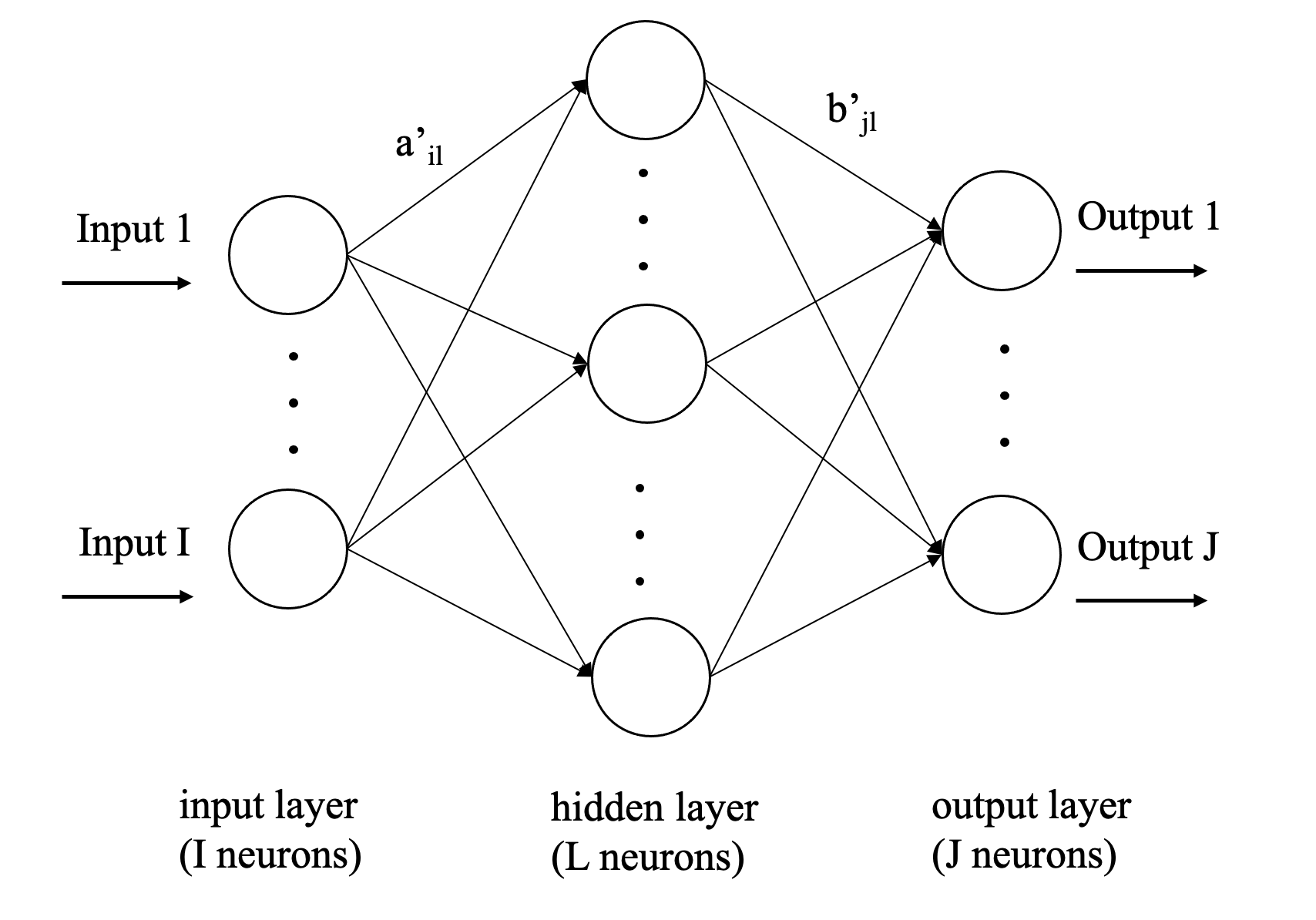} 

}

\caption{The graphical display of LBA-NN}\label{fig:2}
\end{figure}

To improve the ability to predict the response variable, we propose to
extend LBA with an unconstrained feedforward neural network, abbreviated
as LBA-NN. The input and output data matrices in LBA-NN remain the same
as in the preliminary constrained NN extension of LBA. The graphical
display of LBA-NN is shown in Figure \ref{fig:2}. Here, the number of
hidden neurons \(L\) can be larger than \(I\) and/or \(J\), which is
supposed to help catch more patterns in the data and lead to better
prediction.

\par

The model formula of LBA-NN is: \begin{equation}
Y = \sigma_2( \sigma_1 (XA' + b_1)B'^T + b_2) \tag{3.1}
\end{equation} where \(A'\) denotes the \(I \times L\) input-hidden
weight matrix without constraints of constant row sum and
non-negativity; \(B'\) denotes the transpose of \(L \times J\)
hidden-output one without corresponding constraints; \(b_1\) and \(b_2\)
represent the bias term in the hidden layer and output layer,
respectively; \(\sigma_1\) and \(\sigma_2\) are the activation functions
for the hidden layer and output layer, respectively. The neurons in the
input layer are \(I\) categories of the explanatory variable(s).
Likewise, the neurons in the output layer represent \(J\) categories of
the response variable. In contrast, the hidden layer does no longer
directly represent \(K\) latent budgets.

The neurons in a layer can be activated by different functions, linear
(\(\sigma(x)=cx\), \(c\) is the slope), Rectified Linear Unit(ReLU,
\(\sigma(x)=max(0,x)\)) or softmax
(\(\sigma_j(x) = \frac{e^{x_j}}{\sum_{j=1}^J{e^{x_j}}}\)) activation
\citep{nwankpa2018}. The weights in matrix \(A'\) and \(B'\) of LBA-NN
are optimized by the backpropagation algorithm \citep{Rojas1996}. The
algorithm basically treats the outputs as inputs, sends back the bias
and readjusts the weights to minimize the distance between predicted
outputs and actual outputs. That distance is called loss function. Two
commonly used loss functions are mean square error (see Section
\ref{subsection:quanti} for details) \citep{allen1972} and cross-entropy
\citep{hui2021}. The backpropagation algorithm updates the weights in
matrices \(A'\) and \(B'\) iteratively: \begin{equation}
    w^* = w + r(\frac{\partial Loss}{\partial w}) \tag{3.2}
\end{equation} where \(w\) is the old version of the weight; \(w^*\) is
the updated weight; \(r\) is the learning rate, a tuning parameter that
regulates the magnitude of change in each iteration.

\hypertarget{subsection:function}{%
\subsection{\texorpdfstring{Implementation in R:
\textsf{lbann}}{Implementation in R: }}\label{subsection:function}}

An R function, \textsf{lbann}\textsuperscript{\ref{footnote:lbann}} was
developed for LBA-NN. The R function depends on the packages,
\textsf{keras} \citep{keras}, \textsf{ggplot2} \citep{ggplot2},
\textsf{magrittr} \citep{magrittr}, and \textsf{tensorflow}
\citep{tensorflow}. The Algorithm for \textsf{lbann} is shown in
Algorithm 1. The function is composed of 5 steps. Firstly, a feedforward
NN model is set up according to the given number of input, hidden and
output neurons and the activation functions. Then, the corresponding
weights in matrix \(A'\) and \(B'\) are initialized by sampling from a
normal distribution. Third and afterwards, the weights are optimized by
the backpropagation algorithm based on the chosen loss function and
learning rate. Fourth, the importance plots for \(I\) categories of the
explanatory variable(s) are drawn based on the corresponding importance
table. Lastly, \(K\) clusters are found by the \emph{K}-means clustering
on the importance table and a biplot is yielded (see Section
\ref{subsection:interpret} for details). \bigskip

\newcommand\mycommfont[1]{\footnotesize\ttfamily\textcolor{blue}{#1}}
\SetCommentSty{mycommfont}

\begin{minipage}{0.90\textwidth}
\SetNlSty{textbf}{}{:}
\begin{algorithm}[H]
\SetAlgoLined
  
   \textbf{Step 1:} (Construct a Feedforward Neural Network Model)  
            
            \ \ \ \ $\mathrm{Input \leftarrow layer\_input(n_{input})}$
             \tcp*{explanatory variable(s)} 
            
            \ \ \ \ $\mathrm{Hidden \leftarrow layer\_dense(n_{hidden},\: activation)}$ \tcp*{pre-defined by users}
            
            \ \ \ \ $\mathrm{Output \leftarrow layer\_dense(n_{output},\: activation)}$ \tcp*{response categories}
            \tcc{Each neurons in two adjacent layers are fully connected}
                
     \hspace*{\fill} \\
     \textbf{Step 2:} (Initialization)
     
            \ \ \ \ $\mathrm{A' \leftarrow matrix(nrow = n_{input},\: ncol = n_{hidden})}$
            
            \ \ \ \ $\mathrm{B'^T \leftarrow matrix(nrow = n_{hidden},\: ncol = n_{output})}$
            
            \ \ \ \ $\mathrm{w_{initial} \leftarrow rnorm(mean=0.0, stddev=0.05, seed=seed)}$  \tcc{Initialized weights from a normal distribution}      

     \hspace*{\fill} \\
     \textbf{Step 3:} (Backpropagation)
        
        \ \ \ \ \textbf{for} $\mathrm{i \leftarrow 1 \: to \: n_{epochs}:}$ 
        
        \ \ \ \ \ \ \ \ $\mathrm{w_{i} \leftarrow w_{i - 1} + r(\frac{\partial Loss}{\partial w_{i - 1}})}$ \tcp*{update the weights in gradient descent}

      \hspace*{\fill} \\
      \textbf{Step 4:} (Connection Weight Approach)
            
        \ \ \ \ \textbf{for} $\mathrm{i} \leftarrow 1 \: \mathrm{to} \: I$:
        
        \ \ \ \ \ \ \ \ \textbf{for} $\mathrm{j} \leftarrow 1 \: \mathrm{to} \: J$:

        \ \ \ \ \ \ \ \  \ \ $\mathrm{Imp[j, i]} \leftarrow \mathrm{sum(A'[i, ] * B'^T[ ,j])}$
            \tcp*{$J \times I$ importance table}
            
     \hspace*{\fill} \\
     \textbf{Step 5:} (LBA-NN-K-means Approach)
            
        \ \ \ \ \ Assign each point an initialized cluster in {1, ..., K}
        
        \ \ \ \ \ \textbf{While} not converge:
        
        \ \ \ \ \ \ \ \ Reassign points on the closet centroid based on $\underset{k}{\operatorname{argmin}} ||x_i - \mu_k||$
        
        \ \ \ \ \ \ \ \ Update the centroids

\caption{The Algorithm of \textsf{lbann} Function for Latent Budget Analysis - Neural Network (LBA-NN)}
\end{algorithm}
\end{minipage}
\bigskip

The function can be called by:

\texttt{lbann(formula,\ data,\ num.neurons,\ ...)}

The input arguments of \textsf{lbann} are presented in Table
\ref{tab:arguments}. The arguments, ranging from \textsf{num.neurons} to
\textsf{lr}, are the configurations of LBA-NN. The optimal values of
these configurations can be derived through a grid search
\citep{yang1994} based on the package, \textsf{tfruns} \citep{tfruns}.

\begin{table}[H]

\caption{\label{tab:arguments}The arguments of the R function \textsf{lbann}}
\centering
\resizebox{\linewidth}{!}{
\begin{tabular}[t]{l>{\raggedright\arraybackslash}p{20em}ll}
\toprule
\textbf{Arguments} & \textbf{Explanation} & \textbf{Type} & \textbf{Default value}\\
\midrule
formula & An object of class ``formula" (or one that can be coerced to that class) with a response variable. & formula & - \textsuperscript{*}\\
data & A data frame to interpret the variables named in the formula. & data frame & -\\
num.neurons & The number of neurons in the hidden layer. & numeric & -\\
activation.1 & The activation function for the hidden layer, one of ``linear" or ``relu". & character & ``linear"\\
activation.2 & The activation function for the output layer, one of ``linear", ``relu" or ``softmax". & character & ``linear"\\
\addlinespace
loss.function & A loss function refers to an objective function to represent the error. & character & ``mse"\\
epochs & The number of epochs or iterations to train the model. See also ``fit.keras.engine.training.Model" in keras. & numeric & 10\\
val\_split.ratio & The ratio between the number of observations in the training set and validation set. & numeric & 0.2\\
lr & Learning rate. & numeric & 0.01\\
K & The number of predefined clusters. The value is recommended to be the same as in LBA. & numeric & -\\
\addlinespace
seed & The seed for LBA-NN-K-means approach to achieve reproducible results. & numeric & 1\\
\bottomrule
\multicolumn{4}{l}{\textsuperscript{*} ``-": NULL by default}\\
\end{tabular}}
\end{table}

\hypertarget{subsection:interpret}{%
\subsection{Interpretation of the function
results}\label{subsection:interpret}}

To better interpret LBA-NN, a qualitative comparison strategy is
presented in this subsection. The purpose is to find the similarity
between LBA-NN and LBA. LBA was conducted based on the R package,
\textsf{lba} \citep{lba2018}.

Recall that we propose to use a framework of an unconstrained NN in
LBA-NN. However, the NN models have been criticized as ``black boxes''
in the sense that they are statistically unidentifiable
\citep{zhang2018}. Intuitively, in a linear neural network model
(ignoring the bias for the moment) the formula can be simply:
\begin{equation}
  Y = XA'B'^T \tag{4.1}
\end{equation} It is obvious that we can derive many different
combinations of \(A'\) and \(B'\) to get the same predicted \(Y\). In
other words, each time we train the NN model, we are supposed to obtain
different weight matrices, imposing challenges for interpretation. Even
though the weights can vary, the indicator, relative importance of each
level of the explanatory variable(s), is able to produce more stable and
plausible results of the impact on the response variable. The method
adopted to calculate the relative importance is the Connection Weight
Approach \citep{olden2002}, which showed high accuracy when interpreting
NN \citep{olden2004}. The formula for the Connection Weight Approach is:
\begin{equation}
  Imp_{ij} = \sum^L_{l=1}{a'_{il}b'_{jl}} \tag{4.2}
\end{equation} where \(Imp_{ij}\) represents the importance of the
\(i^{th}\) category of the explanatory variable(s) on the \(j^{th}\)
category of the response variable; \(a'_{il}\) is the weight connecting
the \(i^{th}\) category of the explanatory variable(s) and the
\(l^{th}\) hidden neuron; \(b'_{jl}\) is the weight connecting the
\(l^{th}\) hidden neuron and the \(j^{th}\) category of the response
variable. In other words, the relative importance is calculated by
summing the product of weights across the hidden neurons. A
straightforward visualization, importance plot, is provided in
\textsf{lbann}. An importance plot is a bar plot in nature, where the x
axis displays the levels of the explanatory variable(s) and y axis
stands for the importance values. The plot describes the contributions
of all \(I\) categories of the explanatory variable on the \(j^{th}\)
category of the response variable. The more important the explanatory
category is, the higher the bar will be.

After obtaining the relative importance from the categorical explanatory
variable(s) on each level of the response variable, it is easy to form
an importance table where the rows are \(J\) levels of the response
variable, columns are \(I\) categories of the explanatory variable(s),
and cells are the importance values. Each row can be seen as a pattern
showing the contributions from all \(I\) levels of the explanatory
variable(s) on each of the \(J\) categories. To compare LBA-NN and LBA,
\emph{K}-means clustering \citep{hartigan1979} is applied to the
importance table to find \(K\) clusters based on the abovementioned
contribution patterns. We call this approach the LBA-NN-K-means
approach. The \(K\) clusters are comparable to \(K\) latent budgets in
LBA. Lastly, the results of LBA-NN-K-means approach are visualized in a
biplot \citep{gabriel1971}, where \(I\) categories of the explanatory
variable(s) are shown as inverted triangles and \(J\) categories of the
response variable as points.

\hypertarget{subsection:quanti}{%
\subsection{Quantitative evaluation}\label{subsection:quanti}}

In this subsection, we propose to compare LBA-NN and LBA by six standard
measurements of predictive performance, namely mean square error (MSE),
accuracy, precision, recall, specificity, f1-score \citep{lever2016}.
Note that the preliminary constrained NN extension of LBA, formula
\ref{equ:siciliano}, is used here since the traditional LBA does not
have the functionality of predicting the response variable. The mean
square error (MSE) refers to the average similarities between the actual
output vector and predicted output vector. MSE is calculated as:
\begin{equation}
  MSE = \frac{1}{N}\sum_{n=1}^N{(y_n - \hat{y}_n)^2} \tag{5}
\end{equation} where \(y_n\) denotes the observed value of the output,
and \(\hat{y}_n\) is the predicted value. Lower MSE indicates better
prediction. The accuracy is calculated as the proportion of correctly
classified cases among all the observations. The precision is the
average proportion of correctly classified cases among each retrieved
category. The recall is the average proportion of correctly classified
cases among each of the \(J\) categories. The specificity is the average
proportion of cases who do not belong to the \(j^{th}\) category and are
indeed predicted to belong to other categories. The f1-score is
calculated as
\(\frac{2\times precision \times recall}{precision + recall}\),
indicating the balance between precision and recall. Higher accuracy,
precision, recall, specificity or f1-score imply better performance.

\hypertarget{sec:app}{%
\section{Applications}\label{sec:app}}

In this section, we apply the aforementioned R function, \textsf{lbann},
and compare LBA-NN with LBA qualitatively and quantitatively. To give an
insight into the functionality of LBA-NN and \textsf{lbann}, several
experiments are provided. In Section \ref{subsection:example1}, a
simulated dataset with one explanatory variable is analysed. In Section
\ref{subsection:example2}, a freely available
dataset\footnote{\label{footnote:led} \url{http://archive.ics.uci.edu/ml/datasets/LED+Display+Domain}}
with five explanatory variables is used. In addition, we conducted an
extra experiment on previously analysed German Suicide data
\citep{Heijden1992}. The results and discussion of the German Suicide
data are provided in supplementary
materials\footnote{\label{footnote:suppl} \url{https://github.com/ZhenweiYang96/MSc-Statistics-Thesis/blob/main/Supplementary-materials.pdf}}.

\hypertarget{subsection:example1}{%
\subsection{Simulated data with one explanatory
variable}\label{subsection:example1}}

Example 1 is a simulated dataset, whose data generating mechanism is
shown in Table S1 (see supplementary
materials\textsuperscript{\ref{footnote:suppl}}). The training set has
800 observations and the test set 200 observations. One explanatory
variable is included in the models. The explanatory variable \(P\) has
six categories and the response variable \(Y\) four categories. A
summary of the data in a contingency table is presented in Table
\ref{tab:tabex1}. The optimal hyperparameters include 8 hidden neurons,
the ReLU activation for the hidden layer and the softmax activation for
the output layer, with a learning rate of \(10^{-2}\). Three latent
budgets are predefined in LBA.

\begin{table}[H]

\caption{\label{tab:tabex1}The contingency table for the example data 1}
\centering
\begin{tabular}[t]{>{}cccccc}
\toprule
\multicolumn{1}{c}{\textbf{ }} & \multicolumn{1}{c}{\textbf{ }} & \multicolumn{4}{c}{\textbf{Y}} \\
\cmidrule(l{3pt}r{3pt}){3-6}
 &  & 1 & 2 & 3 & 4\\
\midrule
 & 1 & 164 & 2 & 0 & 0\\

 & 2 & 77 & 76 & 1 & 0\\

 & 3 & 9 & 154 & 37 & 0\\

 & 4 & 0 & 18 & 125 & 3\\

 & 5 & 0 & 0 & 81 & 85\\

\multirow{-6}{*}{\centering\arraybackslash \textbf{P}} & 6 & 0 & 0 & 6 & 162\\
\bottomrule
\end{tabular}
\end{table}

The estimated LBA-parameters are shown in Table \ref{tab:lbaex1}. In the
first latent budget, the mixing parameters of \(P\) valued 2
(\(\alpha_{k=1|P=2}\)), 3 (\(\alpha_{k=1|P=3}\)) and 4
(\(\alpha_{k=1|P=4}\)) are larger than the corresponding budget
proportion. The latent budget parameters of \(Y\) valued 2
(\(\beta_{Y=2|k=1}\)) and 3 (\(\beta_{Y=3|k=1}\)) are larger than the
corresponding marginal probabilities (\(p_{+j}\)). Thus, the first
latent budget is used more than average by \(P\) valued 2, 3 and 4 and
\(Y\) valued 2 and 3. Likewise, the second latent budget is used more
than average by \(P\) valued 5 and 6 and \(Y\) valued 3 and 4. The third
latent budget consists of \(P\) valued 1 and 2 and \(Y\) valued 1.

\begin{table}[H]

\caption{\label{tab:lbaex1}Parameter Estimates from LBA for Example 1}
\centering
\begin{tabular}[t]{ccccc}
\toprule
\textbf{Mixing parameters} & \textbf{k = 1} & \textbf{k = 2} & \textbf{k = 3} & \textbf{}\\
\midrule
1 & 0.00 & 0.00 & 1.00 & \\
2 & 0.49 & 0.00 & 0.51 & \\
3 & 0.94 & 0.00 & 0.06 & \\
4 & 0.94 & 0.06 & 0.00 & \\
5 & 0.00 & 1.00 & 0.00 & \\
\addlinespace
6 & 0.00 & 1.00 & 0.00 & \\
\textbf{budget propotion} & 0.39 & 0.33 & 0.27 & \\
\textbf{Latent budgets} & \textbf{k = 1} & \textbf{k = 2} & \textbf{k = 3} & $p_{+j}$\\
1 & 0.00 & 0.00 & 0.98 & 0.27\\
2 & 0.61 & 0.00 & 0.02 & 0.24\\
\addlinespace
3 & 0.39 & 0.25 & 0.00 & 0.24\\
4 & 0.00 & 0.75 & 0.00 & 0.25\\
\bottomrule
\end{tabular}
\end{table}

\begin{figure}[H]
\centering
\includegraphics[width=0.7\textwidth]{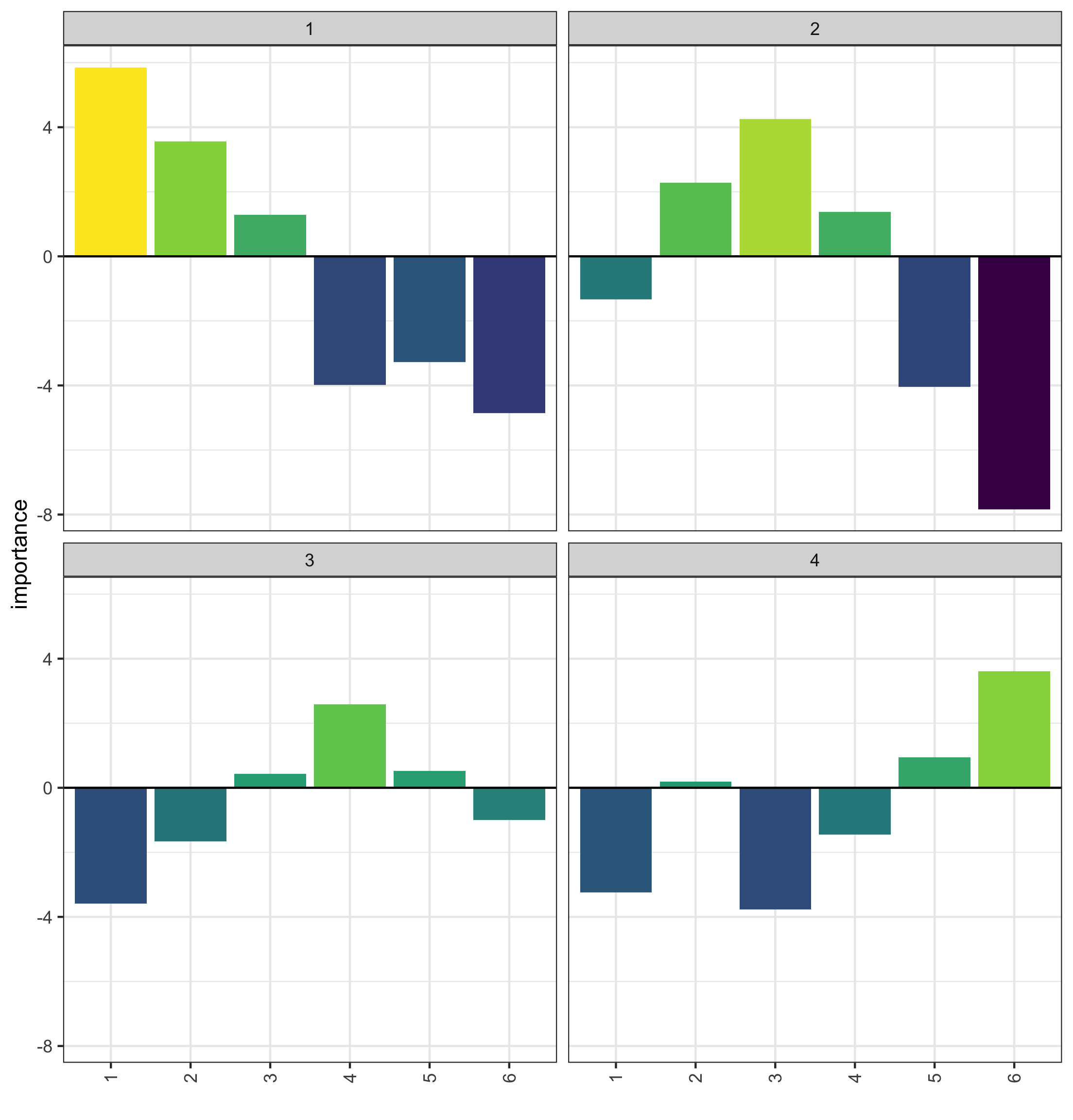}
\caption{The importance plots of example 1 from LBA-NN. (x axis: 6 categroeis of P; y axis: importance value; each grid: Y valued 1, 2, 3 and 4) \label{fig:impex1}}
\end{figure}

\begin{figure}[H]
\centering
\includegraphics[width=0.7\textwidth]{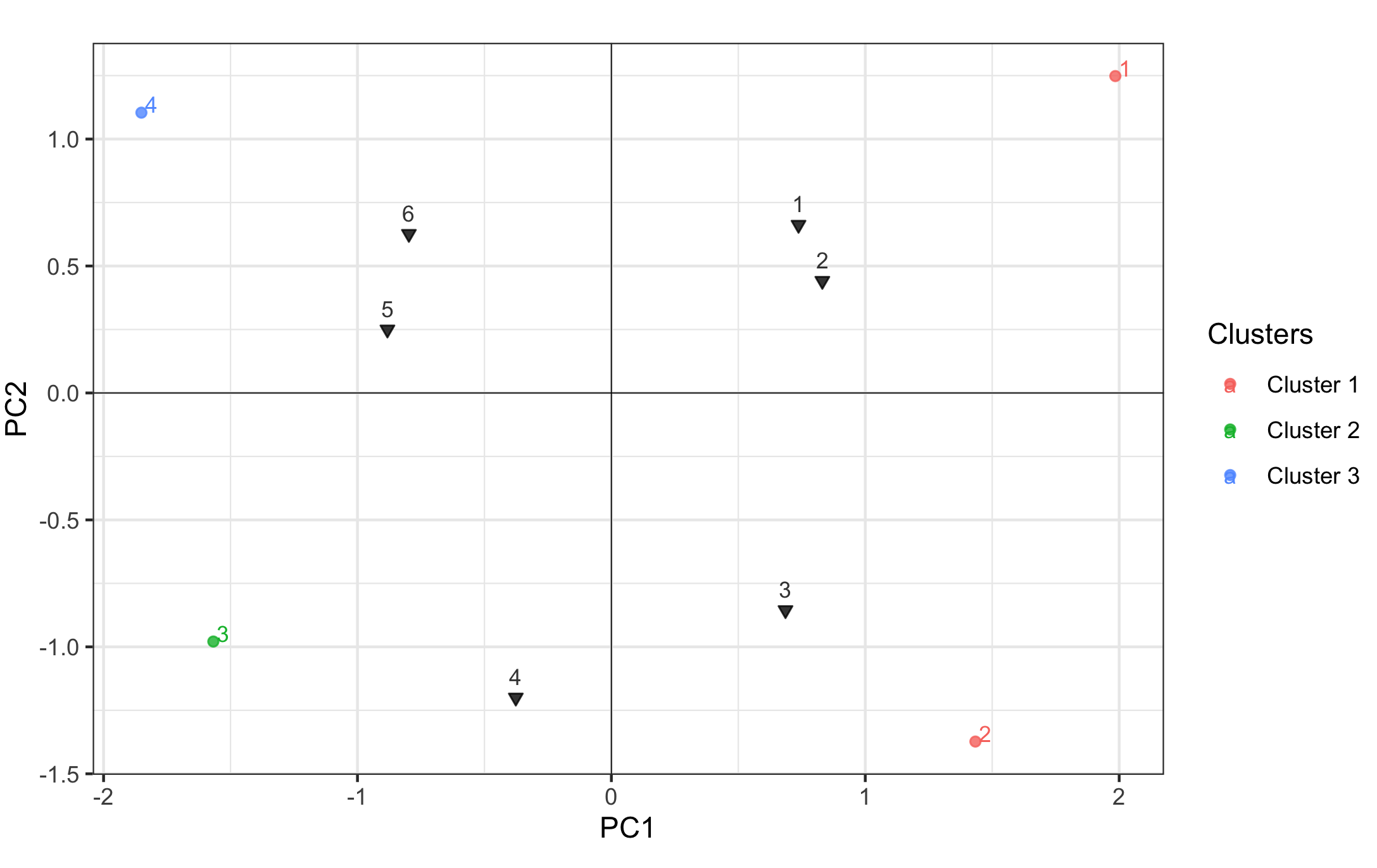}
\caption{The biplot of example 1 from LBA-NN. Three clusters were predefined. The explanatory categories presented in inverted triangle; response variables presented in points.
\label{fig:bipex1}}
\end{figure}

The corresponding qualitative evaluation of LBA-NN is shown in the
importance plots in Figure \ref{fig:impex1}. As in the upper-left panel,
\(P\) valued 1 has the largest impact on \(Y\) valued 1 as the relative
importance, \(Imp_{P=1;Y=1}\), shows the highest value. Similarly, it
can be concluded that \(Y\) valued 2 is mostly related to \(P\) valued 2
and 3; \(Y\) valued 3 is mostly related to \(P\) valued 4; \(Y\) valued
4 is mostly related to \(P\) valued 6. The biplot derived from the
importance table is shown in Figure \ref{fig:bipex1}. Note that three
clusters were predefined so as to compare with LBA (with three latent
budgets) on the same level. The first cluster is characterized by \(P\)
valued 1 to 3 and \(Y\) valued 1 and 2. The second one is mainly for
\(P\) valued 4 and \(Y\) valued 3. The \(P\) valued 5 and 6 and \(Y\)
valued 4 belong to the third cluster.

\begin{figure}[H]
\subfigure[\label{fig:5a}]{\includegraphics[width = 7cm]{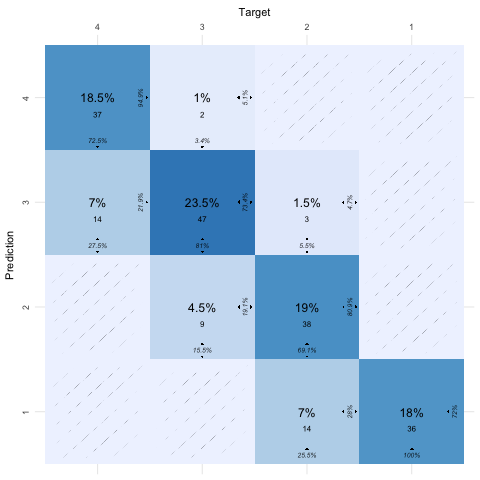}}
\subfigure[\label{fig:5b}]{\includegraphics[width=7cm]{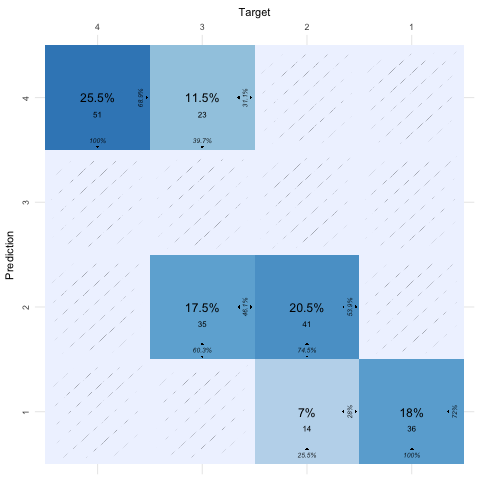}}
\caption{The confusion matrices of LBA-NN and LBA for example 1 (a: LBA-NN; b: LBA). \label{fig:5}}
\end{figure}

Apart from the qualitative evaluation, both models were implemented on
the test data to compare their prediction abilities. The confusion
matrices of the predictions from both models are shown in Figure
\ref{fig:5}. As depicted in Figure \ref{fig:5b}, LBA misclassified \(Y\)
class of 3 to the \(Y\) classes of 2 and 4. The quantitative indicators
are summarized in Table \ref{tab:sumex1}. The accuracy (0.79 versus
0.64), recall (0.81 versus 0.69) and specificity (0.93 versus 0.88) from
LBA-NN are all higher than that from LBA. The mean square error from
LBA-NN is lower than the counterpart from LBA (0.07 versus 0.11). All
indicators imply a better performance in LBA-NN.

\begin{table}[H]

\caption{\label{tab:sumex1}Summary of predicative abilities of the two models for example 1}
\centering
\begin{threeparttable}
\begin{tabular}[t]{>{\raggedright\arraybackslash}p{5cm}>{\centering\arraybackslash}p{3cm}>{\centering\arraybackslash}p{3cm}}
\toprule
\textbf{ } & \textbf{LBA-NN} & \textbf{LBA}\\
\midrule
\textbf{mean square error} & 0.07 & 0.11\\
\textbf{accuracy} & 0.79 & 0.64\\
\textbf{precision} & 0.80 & -\\
\textbf{recall} & 0.81 & 0.69\\
\textbf{specificity} & 0.93 & 0.88\\
\addlinespace
\textbf{f1-score} & 0.80 & -\\
\bottomrule
\end{tabular}
\begin{tablenotes}
\small
\item \textit{Note: } 
\item Recall is also named as sensitivity. Since LBA failed to predict one of the classes (i.e., class 3), the precision and f1-score for LBA are NA.
\end{tablenotes}
\end{threeparttable}
\end{table}

\hypertarget{subsection:example2}{%
\subsection{LED data with five explanatory
variables}\label{subsection:example2}}

Example 2 is a public LED dataset\textsuperscript{\ref{footnote:led}}
from UCI Machine Learning Repository \citep{Dua2019}. The training set
has 400 observations whereas test set has 100 observations. Five
explanatory variables, V1 to V5, are included in the models. All
explanatory variables are binary variables (i.e., with two categories).
A summary of the data is presented in Table \ref{tab:tabex2}. The
optimal hyperparameters include 128 hidden neurons, the linear
activations for the hidden and output layer, learning rate of
\(10^{-4}\). LBA has eight latent budgets.

\begin{table}[H]

\caption{\label{tab:tabex2}The contingency table for the example data 2}
\centering
\begin{threeparttable}
\begin{tabular}[t]{lcccccccccc}
\toprule
\multicolumn{1}{c}{\textbf{ }} & \multicolumn{10}{c}{\textbf{Class}} \\
\cmidrule(l{3pt}r{3pt}){2-11}
  & 1 & 2 & 3 & 4 & 5 & 6 & 7 & 8 & 9 & 10\\
\midrule
V1-0 & 5 & 32 & 8 & 10 & 50 & 4 & 6 & 8 & 8 & 6\\
V1-1 & 40 & 5 & 43 & 47 & 2 & 48 & 41 & 49 & 45 & 43\\
V2-0 & 6 & 36 & 42 & 49 & 4 & 5 & 3 & 54 & 3 & 3\\
V2-1 & 39 & 1 & 9 & 8 & 48 & 47 & 44 & 3 & 50 & 46\\
V3-0 & 2 & 7 & 7 & 2 & 10 & 49 & 42 & 5 & 6 & 3\\
\addlinespace
V3-1 & 43 & 30 & 44 & 55 & 42 & 3 & 5 & 52 & 47 & 46\\
V4-0 & 39 & 35 & 7 & 7 & 4 & 6 & 7 & 52 & 4 & 8\\
V4-1 & 6 & 2 & 44 & 50 & 48 & 46 & 40 & 5 & 49 & 41\\
V5-0 & 4 & 33 & 8 & 55 & 47 & 50 & 5 & 51 & 6 & 43\\
V5-1 & 41 & 4 & 43 & 2 & 5 & 2 & 42 & 6 & 47 & 6\\
\bottomrule
\end{tabular}
\begin{tablenotes}
\small
\item \textit{Note: } 
\item ``V1-0": the LED shows negativity (valued 0) for feature V1; ``V1-1": the LED shows positivity (valued 1) for feature V1, and so on.
\end{tablenotes}
\end{threeparttable}
\end{table}

The estimated parameters from LBA are shown in Table \ref{tab:lbaex2}.
According to the same interpretation strategy, latent budget 1 is mainly
for class 1, 7 and 9 with features of V1, V2 and V5. Latent budget 6 is
composed of LED lights of class 4 and 8 showing positivity in V1 and
negativity in V2 and V5. Notably, in latent budget 8, it can be
concluded that LED lights of class 6 and 7 does not have the feature V3.
LBA does not provide clear information on other features since the
mixing parameters for both of the categories (all mixing parameters
except \(\alpha_{k=8|V3=0}\)) are under the budget proportion.

\begin{table}[H]

\caption{\label{tab:lbaex2}Parameter Estimates from LBA for Example 2}
\centering
\resizebox{\linewidth}{!}{
\begin{tabular}[t]{cccccccccc}
\toprule
\textbf{Mixing parameters} & \textbf{k = 1} & \textbf{k = 2} & \textbf{k = 3} & \textbf{k = 4} & \textbf{k = 5} & \textbf{k = 6} & \textbf{k = 7} & \textbf{k = 8} & \textbf{}\\
\midrule
V1-0 & 0.00 & 0.56 & 0.00 & 0.00 & 0.00 & 0.00 & 0.35 & 0.09 & \\
V1-1 & 0.26 & 0.00 & 0.19 & 0.24 & 0.14 & 0.17 & 0.00 & 0.00 & \\
V2-0 & 0.00 & 0.00 & 0.23 & 0.00 & 0.00 & 0.43 & 0.35 & 0.00 & \\
V2-1 & 0.45 & 0.25 & 0.00 & 0.30 & 0.00 & 0.00 & 0.00 & 0.00 & \\
V3-0 & 0.03 & 0.00 & 0.10 & 0.06 & 0.00 & 0.00 & 0.02 & 0.79 & \\
\addlinespace
V3-1 & 0.00 & 0.22 & 0.28 & 0.05 & 0.45 & 0.00 & 0.00 & 0.00 & \\
V4-0 & 0.00 & 0.00 & 0.00 & 0.00 & 0.72 & 0.00 & 0.20 & 0.08 & \\
V4-1 & 0.14 & 0.25 & 0.28 & 0.27 & 0.00 & 0.06 & 0.00 & 0.00 & \\
V5-0 & 0.00 & 0.24 & 0.00 & 0.28 & 0.00 & 0.30 & 0.18 & 0.00 & \\
V5-1 & 0.72 & 0.00 & 0.25 & 0.00 & 0.02 & 0.00 & 0.00 & 0.01 & \\
\addlinespace
\textbf{budget propotion} & 0.17 & 0.16 & 0.15 & 0.15 & 0.14 & 0.10 & 0.09 & 0.05 & \\
\textbf{Latent budgets} & \textbf{k = 1} & \textbf{k = 2} & \textbf{k = 3} & \textbf{k = 4} & \textbf{k = 5} & \textbf{k = 6} & \textbf{k = 7} & \textbf{k = 8} & $p_{+j}$\\
1 & 0.27 & 0.00 & 0.00 & 0.00 & 0.28 & 0.00 & 0.08 & 0.00 & 0.09\\
2 & 0.00 & 0.05 & 0.00 & 0.00 & 0.14 & 0.00 & 0.51 & 0.07 & 0.07\\
3 & 0.10 & 0.00 & 0.46 & 0.00 & 0.00 & 0.03 & 0.17 & 0.00 & 0.10\\
\addlinespace
4 & 0.00 & 0.18 & 0.21 & 0.00 & 0.09 & 0.45 & 0.00 & 0.00 & 0.12\\
5 & 0.03 & 0.55 & 0.00 & 0.00 & 0.00 & 0.00 & 0.05 & 0.10 & 0.10\\
6 & 0.01 & 0.00 & 0.00 & 0.54 & 0.00 & 0.05 & 0.00 & 0.42 & 0.11\\
7 & 0.26 & 0.00 & 0.09 & 0.12 & 0.00 & 0.00 & 0.00 & 0.37 & 0.10\\
8 & 0.02 & 0.00 & 0.00 & 0.00 & 0.34 & 0.44 & 0.18 & 0.04 & 0.12\\
\addlinespace
9 & 0.28 & 0.10 & 0.24 & 0.03 & 0.06 & 0.00 & 0.00 & 0.00 & 0.11\\
10 & 0.03 & 0.12 & 0.00 & 0.31 & 0.10 & 0.02 & 0.00 & 0.00 & 0.09\\
\bottomrule
\end{tabular}}
\end{table}

\begin{figure}[H]
\centering
\includegraphics[width=0.8\textwidth]{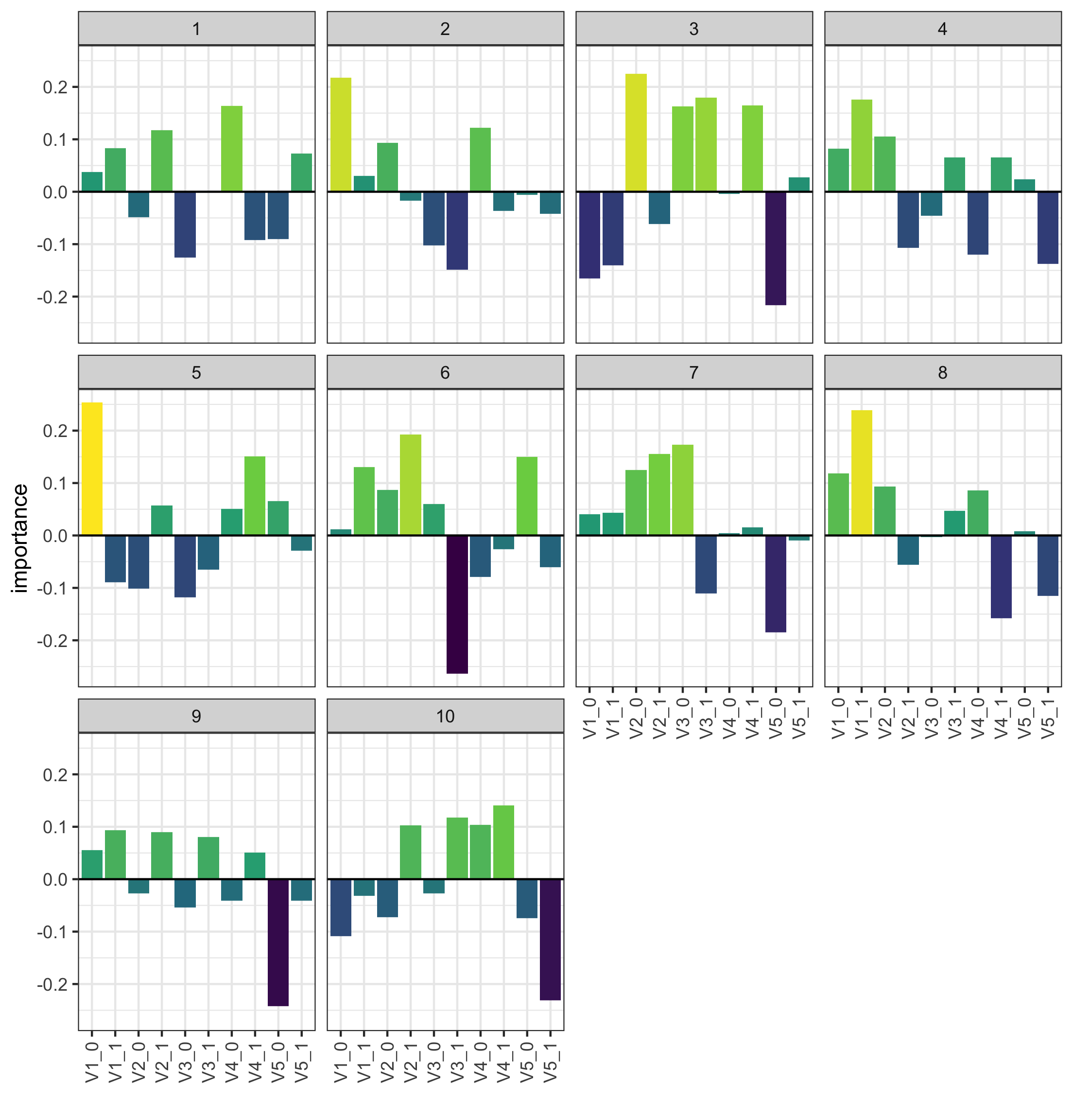}
\caption{The importance plots of example 2 from LBA-NN. (x axis: categories of 5 features; y axis: importance value; each grid: for the categories of the response variables, LED light classes) \label{fig:impex2}}
\end{figure}

\begin{figure}[H]
\centering
\includegraphics[width=0.7\textwidth]{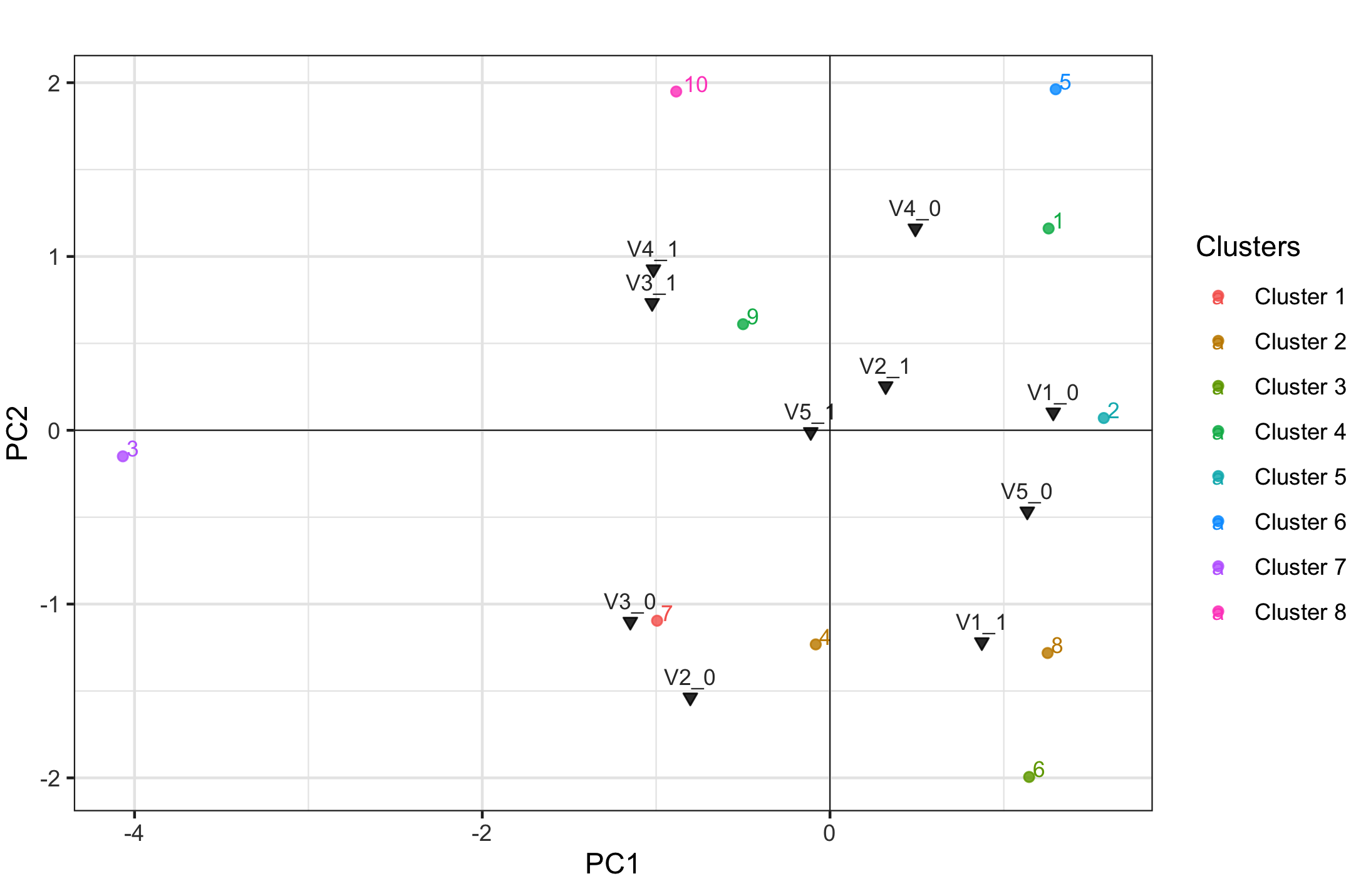}
\caption{The biplot of example 2 from LBA-NN. Eight clusters were predefined. The explanatory categories presented in inverted triangle; response categories presented in points.
\label{fig:bipex2}}
\end{figure}

The corresponding qualitative evaluation of LBA-NN is shown in the
importance plots in Figure \ref{fig:impex2}. For LBA-NN, we interprete
the importance plots by comparing the relative importance of both
categories in the same explanatory variable. As in the first panel, the
class 1 LED lights tend to have all the features except feature V4 as
for all five explanatory variables, only \(Imp_{V4=0;class=1}\) is
greater than \(Imp_{V4=1;class=1}\). While the class 9 LED lights tend
to have all the features ranging from V1 to V5, indicated by the
relative importance of all explanatory variables showing positivity
greater than that of the variables showing negativity
(\(Imp_{Vm=1;class=9} > Imp_{Vm=0;class=9} \text{ for all } m \in \{1, ...,5\}\)).
The biplot derived from the importance table is shown in Figure
\ref{fig:bipex2}. Note that eight clusters were predefined so as to
compare with LBA (with eight latent budgets) on the same level. Cluster
2 is characterized by LED lights of class 4 and 8 and tend to have
feature V1 but not have V2 and V5. The similar interpretation is also
made from latent budget 6 in LBA. The LED lights of class 1 and 9 share
the same attributes of having features V1 and V2. This pattern can be
observed also in latent budget 1 in LBA.

\begin{figure}[H]
\subfigure[\label{fig:8a}]{\includegraphics[width = 7cm]{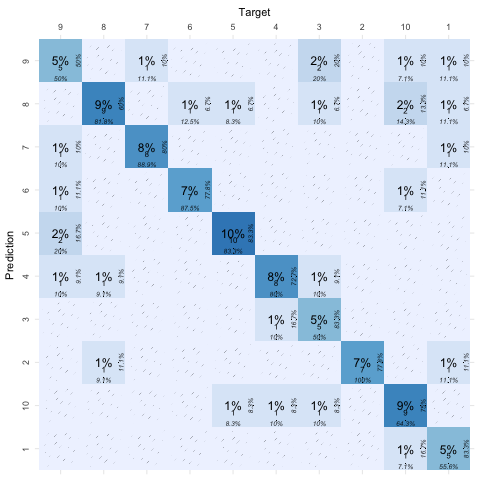}}
\subfigure[\label{fig:8b}]{\includegraphics[width=7cm]{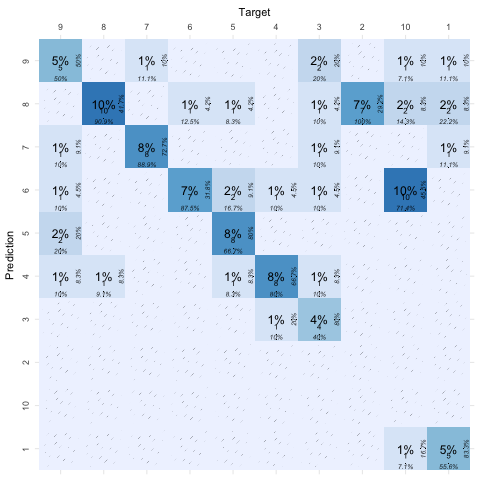}}
\caption{The confusion matrices of LBA-NN and LBA for example 2 (a: LBA-NN; b: LBA). \label{fig:8}}
\end{figure}

Apart from the qualitative evaluation, both models were implemented on
the test data to compare their prediction abilities. The confusion
matrices of the predictions from both models are shown in Figure
\ref{fig:8}. As depicted in Figure \ref{fig:8b}, LBA failed to predict
the LED lights of class 2 and 10. The quantitative indicators are
summarized in Table \ref{tab:sumex2}. The accuracy (0.73 versus 0.55),
recall (0.74 versus 0.56) and specificity (0.97 versus 0.95) from LBA-NN
are all higher than that from LBA. The mean square error from LBA-NN is
lower than the counterpart from LBA (0.06 versus 0.23). All indicators
imply a better performance in LBA-NN.

\begin{table}[H]

\caption{\label{tab:sumex2}Summary of predicative abilities of the two models for example 2}
\centering
\begin{threeparttable}
\begin{tabular}[t]{>{\raggedright\arraybackslash}p{5cm}>{\centering\arraybackslash}p{3cm}>{\centering\arraybackslash}p{3cm}}
\toprule
\textbf{ } & \textbf{LBA-NN} & \textbf{LBA}\\
\midrule
\textbf{mean square error} & 0.06 & 0.23\\
\textbf{accuracy} & 0.73 & 0.55\\
\textbf{precision} & 0.74 & -\\
\textbf{recall} & 0.74 & 0.56\\
\textbf{specificity} & 0.97 & 0.95\\
\addlinespace
\textbf{f1-score} & 0.74 & -\\
\bottomrule
\end{tabular}
\begin{tablenotes}
\small
\item \textit{Note: } 
\item Recall is also named as sensitivity. Since LBA failed to predict one of the classes (i.e., class 2 and 10), the precision and f1-score for LBA are NA. 
\end{tablenotes}
\end{threeparttable}
\end{table}

\hypertarget{sec:dis}{%
\section{Discussion}\label{sec:dis}}

In this study, we propose a new model for the compositional data,
LBA-NN. The model is based on the idea of LBA and the framework of NN.
In addition, we provide an R function, \textsf{lbann}, to establish
LBA-NN with different number of hidden neurons and activation functions.
One of the advantages of this function is that users are allowed to
manually change the configurations to avoid technical issues, such as
overfitting. To obtain the optimal hyperparameters for LBA-NN, it is
also recommended to use grid search. Compared to the package
\textsf{lba} for LBA, \textsf{lbann} is able to yield a model both for
interpretation and prediction. In that sense, if users have basic
knowledge of the explanatory variable(s), LBA-NN is able to predict
which category in the response variable the subject most probably
belongs to.

To answer the second research question, we presented different example
datasets to implement LBA-NN. The results were compared to LBA. In terms
of the quantitative evaluation, LBA-NN has superior performance to the
constrained NN extension of LBA, indicated by lower mean square error,
higher accuracy, recall and specificity. Remarkably, the constrained NN
extension of LBA failed to predict all \(J\) categories of the response
variable in the abovementioned examples. The reason is that all the
input and output neurons are binary nodes valued either zero or one. In
that circumstance, the weights and number of hidden neurons are far more
important since they guide how the data are transited and determine how
many patterns to detect. If the hidden neurons are fewer than the output
neurons, some of the \(J\) categories can be missing in prediction.

Qualitatively, some similarities between LBA-NN and LBA can be observed.
In example 1, both LBA-NN and LBA infer that lower P values are related
to lower Y values and vice versa. The relationship corresponds to the
data generating mechanism. In example 2, cluster 2 composed of LED
lights of class 4 and 8 is similar to latent budget 6 from LBA. The LED
lights of class 1 and 9 are included in cluster 4 from LBA-NN as well as
latent budget 1 from LBA.

In terms of differences between two models, LBA is likely to include the
same category of the response variable in different latent budgets
whereas the clusters in LBA-NN are mutually exclusive with regard to
categories of the response variable. On one hand, the classification in
LBA can enrich the interpretation. On the other hand, it can bring
challenges in decision making, resulting in poor prediction. The main
drawback of LBA-NN is that it cannot yield straightforward parameters,
such as the mixing parameters and latent budget parameters in LBA.
Nevertheless, with the aid of the relative importance, we provide a
stable strategy to interpret LBA-NN.

By answering the two research questions, it is convinced that LBA-NN, in
a nutshell, is able to produce more accurate predictions as well as
similar interpretations to LBA. We propose the potential usage of LBA-NN
in pattern recognition. For example, in the field of medicine, people
can predict which disease they are most likely to have based on
difference symptoms. Another scenario can be in marketing where the
companies are interested in finding the preference of the customers
based on some known features. However, including continuous variables in
LBA-NN is beyond the scope of this study and the current version of the
function. More relevant work is supposed to save for future researches.

\bibliographystyle{tfcad}
\bibliography{reference.bib}

\end{document}